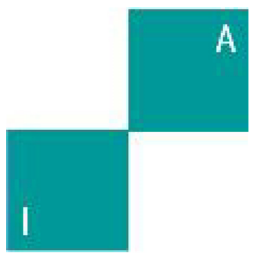

# Breast Cancer Classification Using Gradient Boosting Algorithms Focusing on Reducing the False Negative and SHAP for Explainability

João Manoel Herrera Pinheiro, Marcelo Becker

Mechanical Engineering Department, São Carlos School of Engineering, University of São Paulo, São Carlos 13566-590, SP, Brazil;
joaomh@pm.me; becker@sc.usp.br

**Abstract**: Cancer is one of the diseases that kill the most women in the world, with breast cancer being responsible for the highest number of cancer cases and consequently deaths. However, it can be prevented by early detection and, consequently, early treatment. Any development for detection or perdition this kind of cancer is important for a better healthy life. Many studies focus on a model with high accuracy in cancer prediction, but sometimes accuracy alone may not always be a reliable metric. This study implies an investigative approach to studying the performance of different machine learning algorithms based on boosting to predict breast cancer focusing on the recall metric. Boosting machine learning algorithms has been proven to be an effective tool for detecting medical diseases. The dataset of the University of California, Irvine (UCI) repository has been utilized to train and test the model classifier that contains their attributes. The main objective of this study is to use state-of-the-art boosting algorithms such as AdaBoost, XGBoost, CatBoost and LightGBM to predict and diagnose breast cancer and to find the most effective metric regarding recall, ROC-AUC, and confusion matrix. Furthermore, previous studies have applied Optuna to individual algorithms like XGBoost or LightGBM, but no prior research has collectively examined all four boosting algorithms within a unified Optuna framework, a library for hyperparameter optimization, and the SHAP method to improve the interpretability of our model, which can be used as a support to identify and predict breast cancer. We were able to improve AUC or recall for all the models and reduce the False Negative for AdaBoost and LigthGBM the final AUC were more than 99.41% for all models..

**Keywords**: Breast Cancer, Gradient Boosting Algorithms, Decision Trees, UCI dataset, Optuna, SHAP

## 1 Introduction

Breast cancer is the most common cause of cancer death in women according to the World Health Organization it represents one in four cancer cases and one in six cancer deaths it's the most commonly diagnosed cancer in the world and estimate 2.3 million new cases and 685,000 deaths occurred in 2020 [1, 2]. In Brazil, breast cancer is also one of the most common cancers in women, for each year of the 2023-2025 triennium, 73,610 new cases were estimated [3]. This disease can be classified into multiple subtypes with four widely recognized: luminal A, luminal B, HER2 and triple negative breast cancer (TNBC)[4, 5]. Estimates indicate that the incidence of cancer diagnoses is expected to increase in the coming years, with a projected increase of almost 50% by 2040 compared to 2020 [6]. Detection in time and effective treatment significantly enhance the chance of successful outcomes in breast cancer cases, typically identified by mammography performed by radiologists. The breast comprises three primary components: lobules, connective tissue, and ducts, with cancer that typically originates in the ducts or lobules. Symptoms of breast cancer include lumps or thickening, alterations in size or shape, dimpling, redness, pitting, changes in the appearance of the nipple, and discharge of the nipple. Cancerous tumors represent cells that proliferate abnormally and infiltrate surrounding tissues. Some existing diagnostic methods are mammography, ultrasound, MRI, biopsy, clinical breast examination, and genetic testing. Breast cancer tumors are classified as benign or malignant, with treatment choice based on grade, stage, and molecular subtype of BC, options are surgery, radiation therapy, chemotherapy, neoadjuvant chemotherapy,y and adjuvant chemotherapy[7].

The application of machine learning in the healthcare industry is very common and plays a significant role due to its high performance in prediction, diagnosis, and reduction of the cost of medicine, and breast cancer diagnosis is driven by the aspiration for improved patient outcomes, mitigation of the global impact of the disease, and advancement in healthcare technology and research. Machine learning has demonstrated superior predictive power in various fields, enabling more accurate forecasts and decision-making. For our study, the use of machine learning allows us to achieve more reliable and robust predictions compared to conventional methods, and combining machine learning with advanced optimization tools like Optuna ensures that the models are also fine-tuned to reduce false negatives, which is critical in healthcare.

This study will focus on boosting algorithms, especially in the new state-of-the-art decision tree, like XGBoost, LightGBM, and CatBoost our goal here will be to optimize the recall metric, this metric is very important because it penalizes the false positive in our model, in healthcare it is something that you want to avoid telling someone that they don't have the disease when in true they have. The choice of this gradient boosting model is because for tabular data, they almost have the same performance as other state-of-art algorithms such as Neural Networking, but with a less computational cost.[8, 9].

Our study integrates these boosting algorithms with Optuna, a state-of-the-art hyperparameter optimization framework that leverages Bayesian optimization via Tree-structured Parzen Estimator (TPE). This approach dynamically constructs the search space, enabling more efficient exploration and fine-tuning of hyperparameters tailored to each algorithm. By optimizing for recall, the study aims to reduce false negatives while maintaining strong overall performance as measured by ROC-AUC and other key metrics.

The comprehensive application of Optuna for hyperparameter tuning across four boosting algorithms represents a significant advancement in optimizing predictive performance for breast cancer classification, something not extensively explored in prior studies.

The final goal is to develop a machine learning model with improved performance and without overfitting, however, in some cases it is crucial to find a way to explain the model, especially if some important decision about the result of the model will be taken [10], and in that case, SHAP is very powerful in explaining the black box model. This helps in the final diagnosis by a methodology that can classify the variables in our model[11, 12] but not only that an explained model can help us to find if there had been any data leaks and our model is overfitting [13].

To our knowledge, this study is the first to combine these four boost algorithms with Optuna for hyperparameter optimization. Previous studies have focused on individual algorithms or different optimization techniques. By integrating these advanced techniques, our study not only improves model performance but also enhances interpretability, addressing a critical need in machine learning-driven medical diagnostics.

# 2 Related Work

In recent years, there has been an increase in the use of machine learning techniques in the healthcare area, especially in the detection of breast cancer [14] which usually relies on metrics such as ROC-AUC or accuracy [15, 16].

One notable application is in breast cancer detection, where deep learning methodologies have shown promise in computer-aided diagnosis [17]. For example, in [18], a comprehensive overview was provided for the detection and localization of calcification and breast masses. Additionally, [19] used a combination of supervised (Relief algorithm) and unsupervised (Autoencoder, PCA algorithms) techniques for feature selection, integrated into a Support Vector Machine (SVM) classifier, resulting in accurate and timely detection of breast cancer.

Other researchers have also used the data set mentioned in [20], achieving notable results. For example, in [21], the Random Forest and Support Vector Machine (SVM) models achieved an accuracy of 96.5%. In a comprehensive study conducted by [22], various algorithms including SVM, Naive Bayes (NB), Random Forest (RF), Decision Tree (DT), K-Nearest Neighbor (KNN), Logistic Regression (LR), Multilayer Perceptron (MLP), Linear Discriminant Analysis (LDA), XGBoost (XGB), Ada-Boost, and Gradient Boosting (GBC) were explored. GBC achieved the best accuracy with 99.12%. Additionally, [23] split the data into a set of training 70% and tests 30%, employing six classification models: Linear Support Vector Classification (SVC), Support Vector Classification (SVC), K-Nearest Neighbor (KNN), Decision Tree (DT), Random Forest (RF), and Logistic Regression (LR). Random Forest obtained the highest accuracy of 96.49%, while comparing various performance metrics such as accuracy, AUC, precision, recall, and F1-score. Furthermore, in [24], SVM, Random Forest, Logistic Regression, Decision tree (C4.5), and K-Nearest Neighbors (KNN) were compared based on metrics including AUC, precision, sensitivity, accuracy, and F-Measure. SVM achieved a higher efficiency of 97.2% in the AUC metric. Furthermore, [25] reported a higher accuracy using SVM, with a training accuracy of 99.68%. Their findings contribute to research exploring machine learning techniques for healthcare applications.

In the study by [26], various algorithms including Support Vector Machine, K-Nearest Neighbour, Naive Bayes, Decision Tree, K-means, and Artificial Neural Networks were evaluated, with Artificial Neural Networks achieving the highest accuracy of 97.85%. On the other hand, [27] employed Deep Learning with Adamâs optimization for classification, resulting in a final system accuracy of 96%. In the research conducted by [28], Logistic Regression, Support Vector Machine, Random Forest, Decision Tree, and AdaBoost were utilized, with an impressive accuracy of 98.57%. Furthermore, [29] divided the dataset into a training/test split of 0.67 and 0.33, employing the Random Forest and XGBoost algorithms. The best performance was attained by Random Forest with an accuracy of 74.73%. Lastly, [30] explored classification and clustering algorithms, with decision tree and Support Vector Machine models demonstrating the best performance, achieving an accuracy of 81%. However, the clustering algorithm only achieved 68% accuracy. In [31] they study the Wisconsin Breast Cancer (Original) data set [32], they used min-max normalization for feature scaling and built classifiers using K-nearest neighbor (KNN), SVM, and Logistic Regression. Their results showed a training accuracy ranging from 93% to 97%.

In another study referenced by [33], algorithms such as C4.5, Support Vector Machine (SVM), Naive Bayes (NB), and K-Nearest Neighbor (KNN) were used. Evaluation metrics such as precision and recall were considered, and SVM achieved an accuracy of 97.13%. In addition, [34] introduced an ensemble voting system combining the predictions of five machine learning models, such as Random Forest, Naive Bayes, and SVM. This ensemble system achieved an impressive accuracy of 99.28%.

The researchers mentioned in [35] collected data from three hospitals in Bangladesh and used five machine learning algorithms: Decision tree, Random Forest, Logistic Regression, Naive Bayes, and XGBoost. Their study reported an accuracy of 94% for both Random Forest (RF) and XGBoost. Lastly, in the study described in [36], a hybrid approach was proposed that combines the genetic algorithm and the k-nearest neighbor (KNN). This approach produced an accuracy of 99%.

# 3 Methodology

For this study, a flow chart presenting the phases of this draw in Fig.1 the data were extracted from Breast Cancer Wisconsin [20] which has 569 instances with two classes: Benign (357) and Malignant (212) as shown in Fig.2. Furthermore, the data set contains 31 attributes that were used for analysis and modeling in this study. A Spearman's correlation plot was made (Fig.3). And finally, a table was made that contains all attributes, types, and values range in Table1. Hyperparameter optimization was performed using Optuna, enabling efficient exploration of a wide search space with dynamic pruning based on early performance metrics.

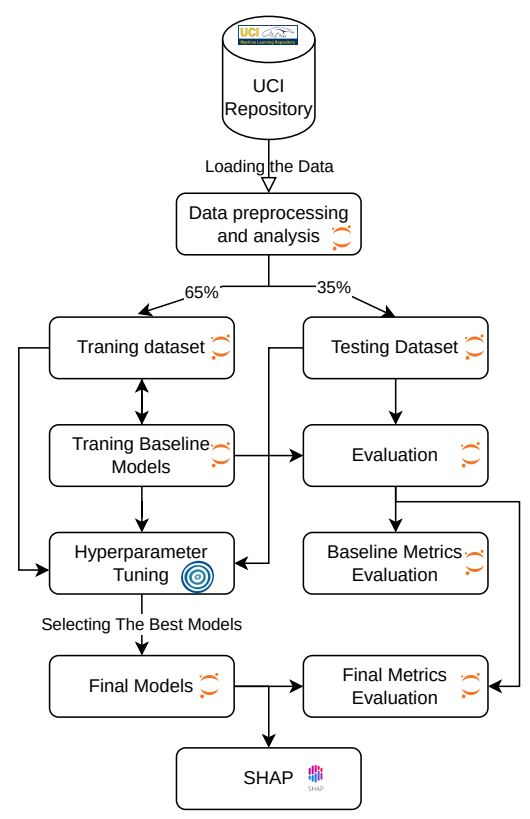

Figure 1: A flow chart presenting the phases of the baseline and final models.

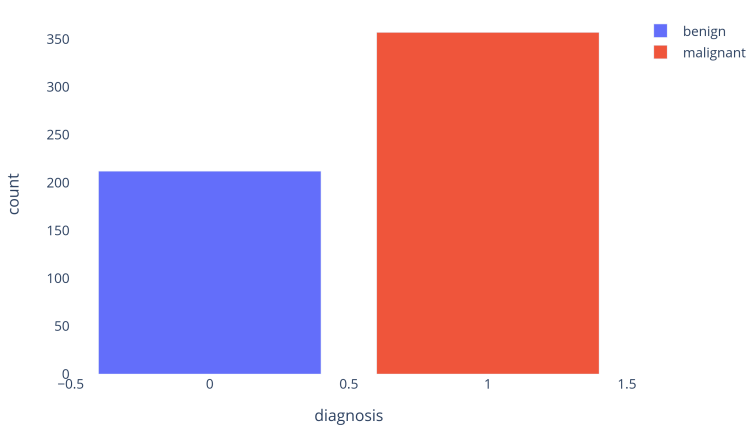

Figure 2: Wisconsin Breast Cancer Target Distribution

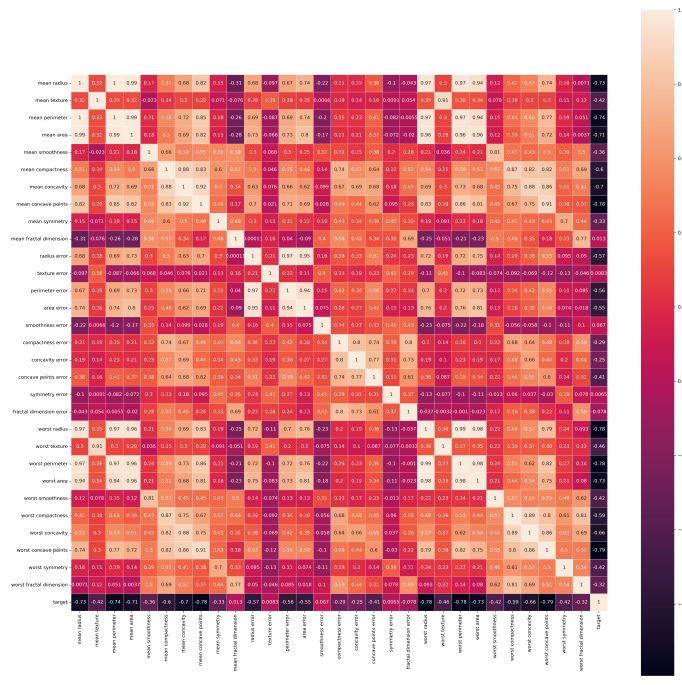

Figure 3: Spearman's correlation

Table 1: Breast Cancer Dataset

| **Attributes** | **Types** | ***Values*** |
|---|---|---|
| mean radius | float64 | 6.981 to 28.11 |
| mean texture | float64 | 9.71 to 39.28 |
| mean perimeter | float64 | 43.79 to 188.5 |
| mean area | float64 | 143.5 to 2501 |
| mean smoothness | float64 | 0.053 to 0.163 |
| mean compactness | float64 | 0.019 to 0.345 |
| mean concavity | float64 | 0 to 0.427 |
| mean concave points | float64 | 0 to 0.201 |
| mean symmetry | float64 | 0.106 to 0.304 |
| mean fractal dimension | float64 | 0.05 to 0.097 |
| radius error | float64 | 0.112 to 2.873 |
| texture error | float64 | 0.36 to 4.885 |
| perimeter error | float64 | 0.757 to 21.98 |
| area error | float64 | 6.802 to 542.2 |
| smoothness error | float64 | 0.002 to 0.031 |
| compactness error | float64 | 0.002 to 0.135 |
| concavity error | float64 | 0 to 0.396 |
| concave points error | float64 | 0 to 0.053 |
| symmetry error | float64 | 0.008 to 0.079 |
| fractal dimension error | float64 | 0.001 to 0.03 |
| worst radius | float64 | 7.93 to 36.04 |
| worst texture | float64 | 12.02 to 49.54 |
| worst perimeter | float64 | 50.41 to 251.2 |
| worst area | float64 | 185.2 to 4254 |
| worst smoothness | float64 | 0.071 to 0.223 |
| worst compactness | float64 | 0.027 to 1.058 |
| worst concavity | float64 | 0 to 1.252 |
| worst concave points | float64 | 0 to 0.291 |
| worst symmetry | float64 | 0.157 to 0.664 |
| worst fractal dimension | float64 | 0.055 to 0.207 |
| target | int | 0 or 1 |

## 3.1 Datapreprocessing

Data preprocessing is an important part of every machine learning project. Normalization is a technique to transform features to be on a similar scale, some models such as Neural networks, Support Vector Machines (SVM), and K-nearest neighbors (KNN) need these transformations. But decision tree-based models, like gradient boost machines used in this study, are generally robust to the scale of features and are not influenced by linear transformations like normalization as much as other models. Decision trees make splits based on thresholds for individual features, and these splits are not affected by linear transformations[37, 38, 39].

Normalization techniques may not lead to significant performance improvements for the models used in this study, as the data consists of numerical values and lacks any missing values, as demonstrated in 1. As a result, other preprocessing steps such as handling missing values and encoding categorical variables become unnecessary.

## 3.2 Gradient Boosting Machine

Boosting is one of the most powerful learning concepts introduced in recent decades. The idea behind boosting is to combine the outputs of many "weak" classifiers to produce a powerful model, essentially creating a "strong" learner from a combination of "weak" learners [40, 41]. Gradient Boosting Machines (GBMs) are a combination of additive models with gradient descent, the GBMs optimize an loss function using the gradient descent. But this optimization is not performed in terms of numerical, instead, they are optimized by boosting functions in the direction of the gradient[42].

$$(\beta_m, \alpha_m) = \arg\min_{\beta,\alpha} \sum_{i=1}^{n} L\left( y^{(i)}, F_{m-1}\mathbf{x}^{(i)} + \beta h(\mathbf{x}^{(i)}; \alpha) \right) \quad (1)$$

Using vectored notation:

$$F_m(\mathbf{X}) = F_{m-1}(\mathbf{X}) + \eta\Delta_m(X) \quad (2)$$

Adaboost, XGBoost, CatBoost, and LightGBM are all based on this boosting theory, they used decision tree as their weak classifier and then some technique to improve the model in each interation.

### 3.2.1 AdaBoost

Adaptive Boosting (AdaBoost) was the first to generate a strong classifier from a set of weak classifiers. The AdaBoost algorithm generates a series of weak learners by managing a set of weights assigned to the training data, adjusting them adaptively after each weak learning iteration [43].

$$G(x_i) = \alpha_1 G_1(x_i) + \alpha_2 G_2(x_i) + ... + \alpha_k G_k(x_i) \quad (3)$$

$$G(x) = sign(\sum_{m=1}^{M} \alpha_m G_m(x)) \quad (4)$$

In which $\alpha_1, \alpha_2, \alpha_M$ are calculated by the boosting algorithm, and each contribution follows a weight in $G_m(x)$ [43].

### 3.2.2 XGBoost

was released in 2016, and is a highly scalable, flexible, and versatile solution, designed to properly exploit features and overcome the limitations of previous gradient boosting. The main difference between XGBoost and other gradient algorithms is that it uses a novel regularization technique to control overfitting[44]. Therefore, it is faster and more robust during model fitting. The regularization technique is accomplished by adding a new term to the loss function, such as

$$L(\phi) = \sum_{i=1}^{n} L(\hat{y}_i, y_i) + \sum_{m=1}^{M} \Omega(f_k) \quad (5)$$

$$\Omega(f) = \gamma T + \frac{1}{2}\lambda||w||^2 \quad (6)$$

### 3.2.3 CatBoost

was develop by Yandex researcher and uses target-based statistics to avoid overfitting, only considers previous data points to calculate the average value, avoiding data leakage, and its main feature is a built-in permutation technique for dealing with categorical columns, one_hot_max_size (OHMS). CatBoost addresses the exponential growth of feature combinations using the greedy method at each new split of the current tree. Furthermore, the construction is performed using symmetric trees and the algorithm supports categorical columns [45].

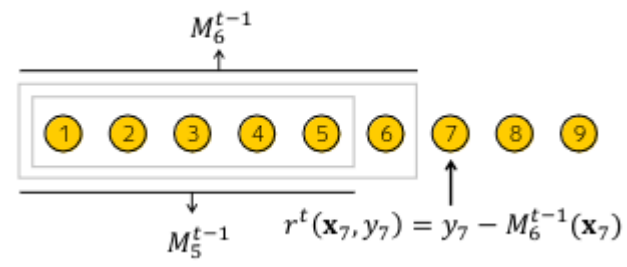

Figure 4: Ordered boosting principle [45]

### 3.2.4 LightGBM

LightGBM is similar to XGBoost with some alterations: a new technique for estimating information gain called gradient-based one-side sampling (GOSS). Since one of the most time-consuming tasks in the gradient boosting learning process is finding the split of the trees, usually some sort of sampling is performed at this stage for efficiency purposes. With that LGBM employs a leaf-wise tree growth strategy as opposed to level-wise, reducing the number of nodes and improving computational efficiency, and then creating an asymmetric tree. It also provides support for categorical features [46].

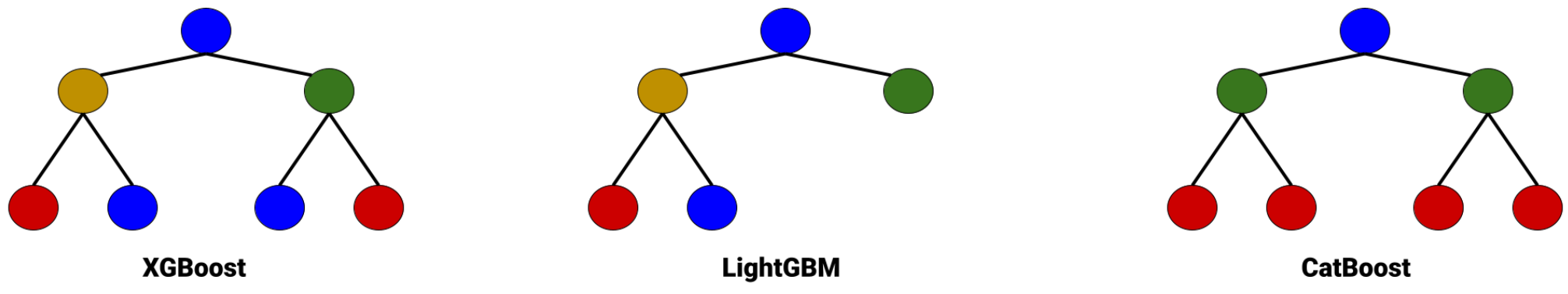

Figure 5: Comparison of Tree Growth in XGBoost, CatBoost, and LightGBM [15]

## 3.3 Metrics of Evaluation

### 3.3.1 Confusion Matrix

One of the metrics for the evaluation of a classification algorithm is a confusion matrix whose table shows the frequency of all classifications with labels such as: True positive (TP), false positive (FP), false negative (FN) and false negative (TN). They are calculated by organizing all data in the confusion matrix, depending on the predicted class by our model and the actual class of each instance in the dataset.

Table 2: Confusion matrix for the binary classification problem.

| **Predicted Class** | **Actual Class** | |
|---|---|---|
| | **Negative** | **Positive** |
| **Negative** | True Negative (TN) | False Positive (FP) |
| **Positive** | False Negative (FN) | True Positive(TP) |

### 3.3.2 ROC-AUC

The Receiver Operating Characteristic (ROC) curve is a probability curve that illustrates the True Positive Rate (TPR) against the False Positive Rate (FPR) across different threshold values. The Area Under the Curve ROC (AUC) quantifies the classifier's capability to discriminate between classes. As discussed in [47, 48, 49], the ROC curve is calculated by plotting the True Positive Rate (TPR), or Sensitivity, against the False Positive Rate (FPR), or Specificity, where $\mathcal{T}$ represents the threshold.

$Roc_y(\mathcal{T}) = TPR_{\mathcal{T}} = \frac{TP_{\mathcal{T}}}{TP_{\mathcal{T}}+FN_{\mathcal{T}}}$

$Roc_x(\mathcal{T}) = FPR_{\mathcal{T}} = \frac{FP_{\mathcal{T}}}{FP_{\mathcal{T}}+TN_{\mathcal{T}}}$

Figure 6: Two points are highlighted on the ROC curve of a classifier to show different values of Specificity and Sensitivity.

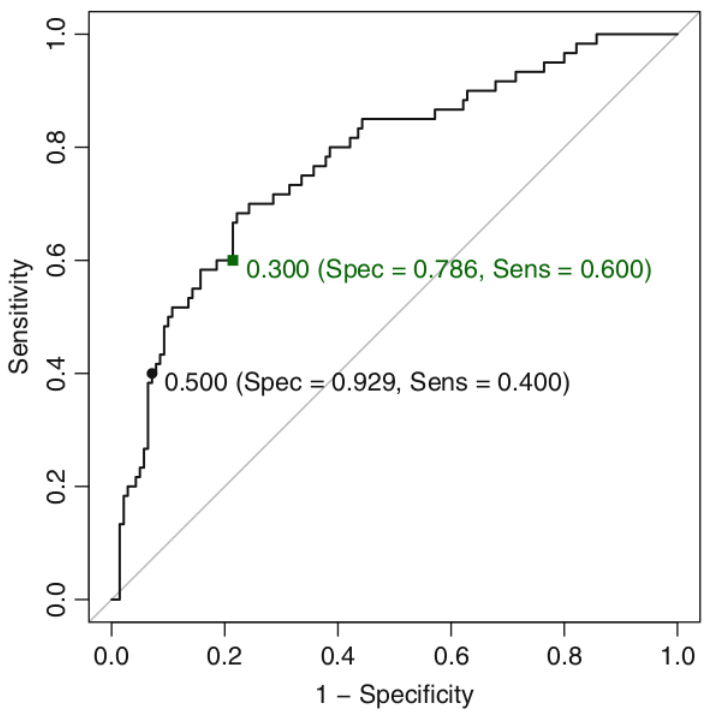

The AUC is defined as the area under the ROC curve and provides a measure of the classifier's performance across all possible thresholds, the value ranges from 0.5 to 1.

### 3.3.3 Accuracy

The number of correct predictions divided by the total number of input samples.

$$Accuracy = \frac{TP + TN}{TP + FP + TN + FN} \tag{7}$$

As discussed in 2, many studies only focus on accuracy when dealing with machine learning models, such as the Breast Cancer Wisconsin (Diagnostic) dataset. However, this metric is sensitive to the effects of class imbalance. Consequently, the model can predict the majority class (class 0) more frequently than the minority class (class 1). In such scenarios, although the model may show high accuracy, it could still produce a considerable number of false negatives. In diagnostic applications of diseases, it is crucial to prioritize the reduction of false negatives.

### 3.3.4 Recall

The number of correct positive results is divided by the number of all relevant samples, true positive plus false positive.

$$\frac{TP}{TP + FN} \tag{8}$$

### 3.3.5 Precision

The number of correct positive results is divided by the number of all relevant samples, true positive plus false positive.

$$\frac{TP}{TP + FP} \tag{9}$$

#### 3.3.6 $F_{measure}$

Is metric is calculated from precision and recall. There exists a trade-off in precision recall metrics, when you increase recall, for example, it will reduce precision and vice versa. A more general $F_{measure}$ can be written as weighted $F_{measure}$ or $F_\beta$[50] .

$$F_\beta = \frac{1+\beta^2}{\frac{\beta^2}{recall} + \frac{1}{precision}} \tag{10}$$

We can write F-M as

$$F_\beta = (1+\beta^2)\frac{precision \times recall}{(\beta^2 \times precision) + recall} \tag{11}$$

Choosing $\beta$ determines the ratio by which recall is weighted higher than precision and it's an empirical metric [51, 52][53].

When $\beta = 1$ the $F_\beta$ is simply a harmonic mean of precision and recall also knowing as $F_1$ or $Fscore$.

$$F_1 = \frac{2 \times precision \times recall}{precision + recall} \tag{12}$$

### 3.4 Training Data

The data was divided into 65% for the training and 35% for evaluation metrics in the test using train_test_split from sklearn.

The 65%-35% split was selected considering the relatively small size of the dataset, ensuring sufficient data for both training and testing phases. Furthermore, this split aligns with similar studies, allowing for more meaningful comparisons of results [21, 22, 23, 24].

### 3.5 Tuning the model

The model was tuned using Optuna, an open source library that allows us to dynamically construct the parameter search space. It is based on Bayesian optimization, more specifically employing a tree-structured Parzen estimator (TPE)[54]. Each study in which the model is trained is in our $X_{train}$ and the metrics are calculated in $X_{test}$.

A framework was developed to evaluate various boosting algorithms using Optuna for hyperparameter optimization, the $F_\beta$ score serving as the key performance metric in the optimization process. The framework is designed for flexible experimentation, allowing comprehensive testing and comparison across different boosting techniques while fine-tuning parameters to achieve optimal performance.

### 3.6 Explainability using SHAP

In some cases, it is extremely important to bring the explainability of the model to help with decision making. For this, SHAP (SHapley Additive exPlanations) finds a way out specific by calculating the contribution of each attribute of the input to the prediction it gives and the individual importance for each instance, it was based on Shapley values from game theory [55, 56]

$$\phi_i(f, x') = \sum_{z' \subseteq x'} \frac{|z'|!(M - |z'| - 1)!}{M!}[f_x(z') - f_x(z' \setminus i)] \tag{13}$$

## 4 RESULTS & Discussion

Throughout the research 8 models were evaluated, first, we created a baseline model and then ran Optuna for each model trying to optimize the $F_\beta$ but also evaluating other metrics such as AUC and Accuracy. The final code can be found in GitHub.

Figure 7: Example of SHAP showing the importance of each input variable on the model output [57].

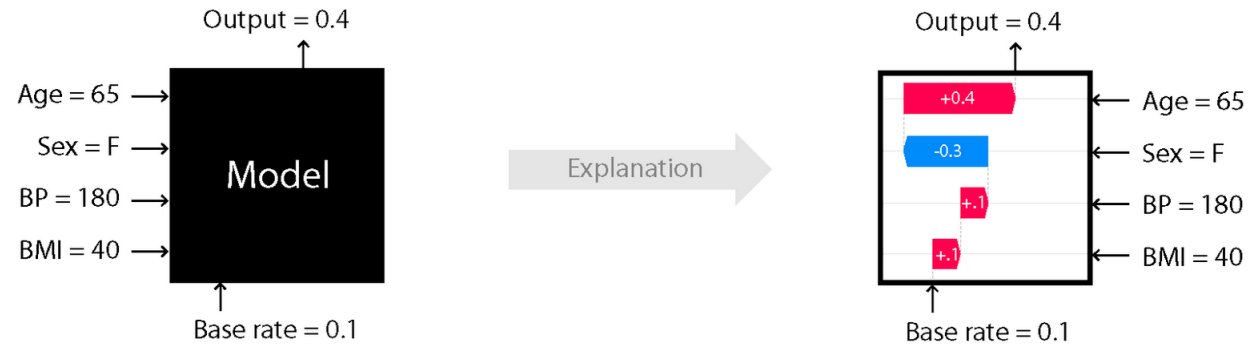

Figure 8: SHAP assigning each feature an importance value[58]

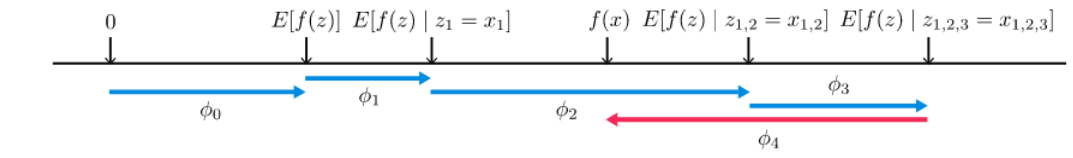

.

The first step was to load the data into DataFrames using pandas and the function describe() to analyze the information in the dataset, max, min, mean and median values, with that information we were able to construct table 1. Then we use the corr() function to plot the Spearman's correlation in Fig 3 using the seaborn library.

## 4.1 Baseline Model

The first thing was to create our baseline models for AdaBosst, XGBoost, CatBoost and LightGBM without tuning any of the hyperparameters, we train at our $X_{train}$ and evaluate the metrics in $X_{test}$. The metrics we evaluated were AUC, Recall, Accuracy, and F1-Score for these baseline models. Here, we have Table 3 with these results.

Then we created the confusion matrix and plotted the ROC curve for all models.

Table 3: Baseline Models Results

| **Model** | **AUC** | **Recall** | **Precision** | **Accuracy** | **F1-Score** |
|---|---|---|---|---|---|
| AdaBoost | 0.9968 | 0.969 | 0.9718 | 0.97 | 0.9766 |
| XGBoost | 0.9932 | 0.9845 | 0.9154 | 0.96 | 0.9695 |
| CatBoost | 0.9959 | 0.9922 | 0.9436 | 0.975 | 0.9808 |
| LightGBM | 0.9922 | 0.9612 | 0.9295 | 0.95 | 0.9612 |

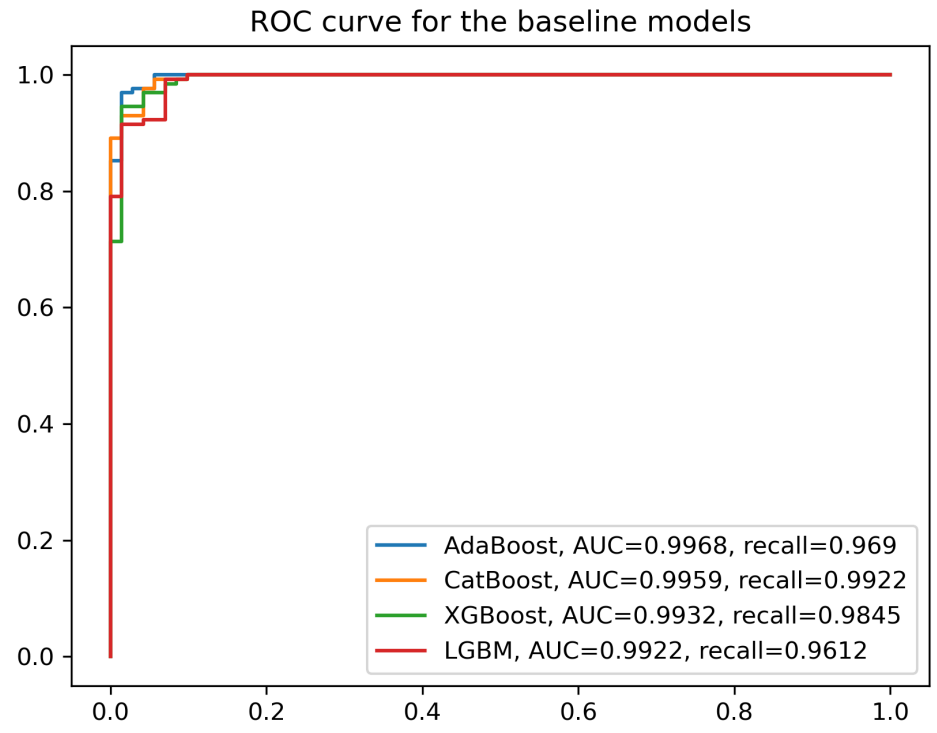

Figure 9: ROC Curve for the four baseline models

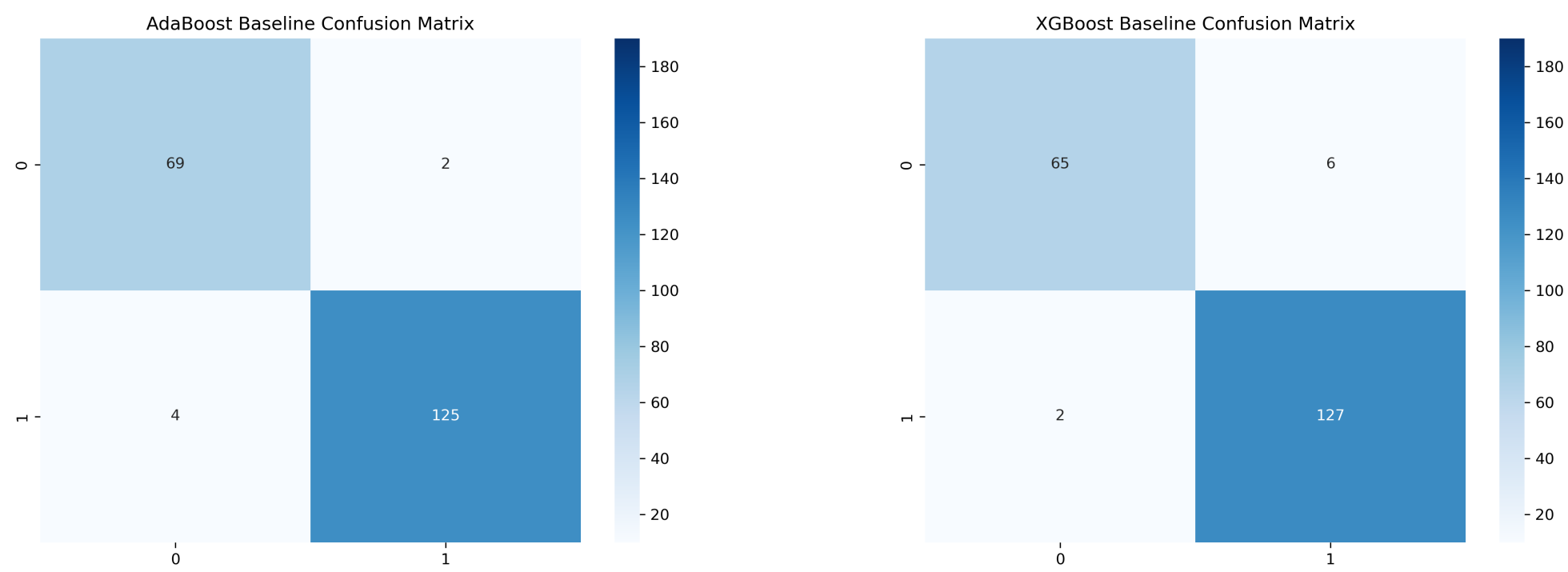

Figure 10: Confusion Matrix for AdaBoost and XGBoost

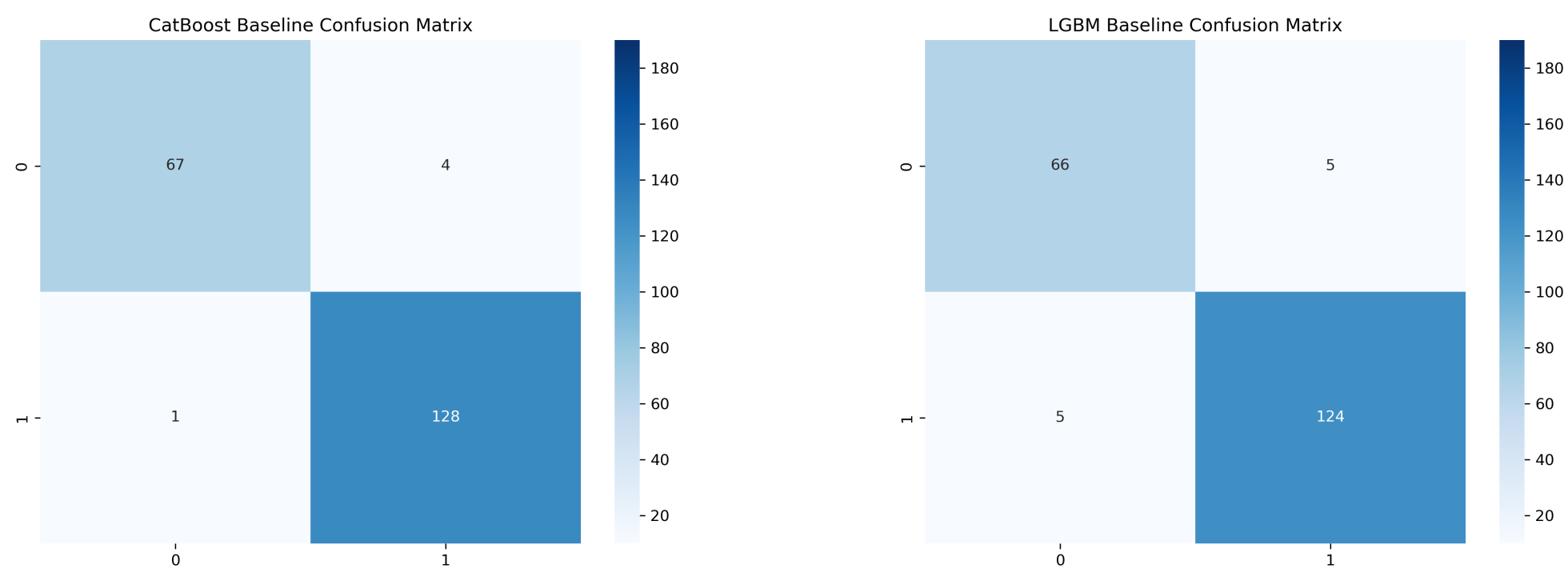

Figure 11: Confusion Matrix for CatBoost and LGBM

## 4.2 Tuning Model Optuna

An objective function was formulated, incorporating the $F_\beta$ metric, to maximize it. This prioritized improving recall by penalizing false negatives, aiming to reduce them compared to our baseline models. Furthermore, a hyperparameter space search was set for each trial in our study and, through several empirical tests, our best results were achieved with $beta = 2.7$.

In the final study in Optuna, AdaBoost conducts 3074 trials, resulting in the best model achieving an $F_\beta$ of 97.58%. XGBoost, after 2830 trials, exhibited its best performance in 2,274 trials, achieving a $F_\beta$ score of 99.13%. CatBoost, subjected to 838 trials, performed better in 102 trials, achieving a $F_\beta$ score of 99.04%. LightGBM, in 3018 trials, demonstrated its highest performance in 127 trials, obtaining a $F_\beta$ score of 99.26%. In Fig 15 we have the final ROC curve for all models tuned by Optuna and their AUC and Recall metrics. The final confusion matrix obtained is shown in Fig 16 and Fig 17.

Finally, we run SHAP for the XGBoost, CatBoost and LightGBM models, AdaBoost doesn't have support for SHAP values natively without changing the source code of SHAP, the results are shown in Fig 4.2.

Table 4: Final Models Results

| **Model** | **AUC** | **Recall** | **Precision** | **Accuracy** | **F1-Score** | $\mathbf{F}_\beta$ |
|---|---|---|---|---|---|---|
| AdaBoost | 0.9941 | 0.9767 | 0.9436 | 0.965 | 0.973 | 0.9758 |
| XGBoost | 0.9956 | 0.969 | 0.9577 | 0.985 | 0.9884 | 0.9913 |
| CatBoost | 0.9967 | 0.9952 | 0.9436 | 0.98 | 0.9884 | 0.9904 |
| LightGBM | 0.9949 | 1.0 | 0.8873 | 0.96 | 0.9699 | 0.9926 |

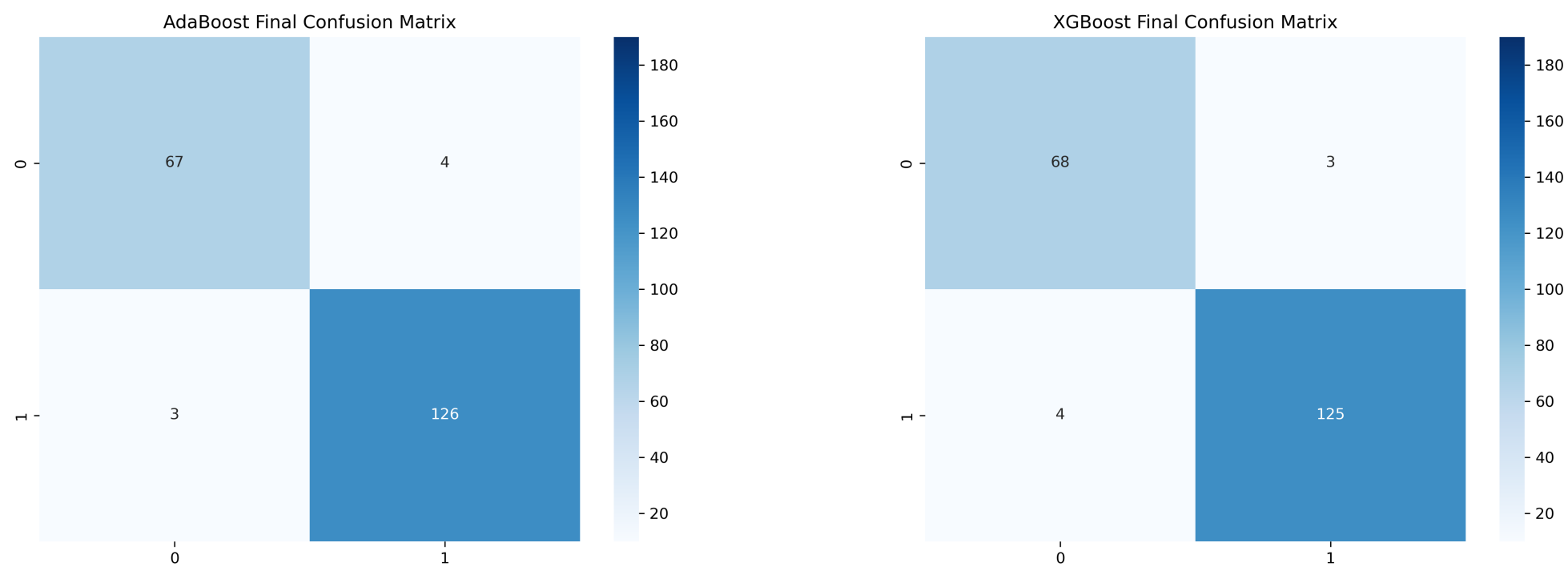

Figure 12: Confusion Matrix for AdaBoost and XGBoost - Final Model

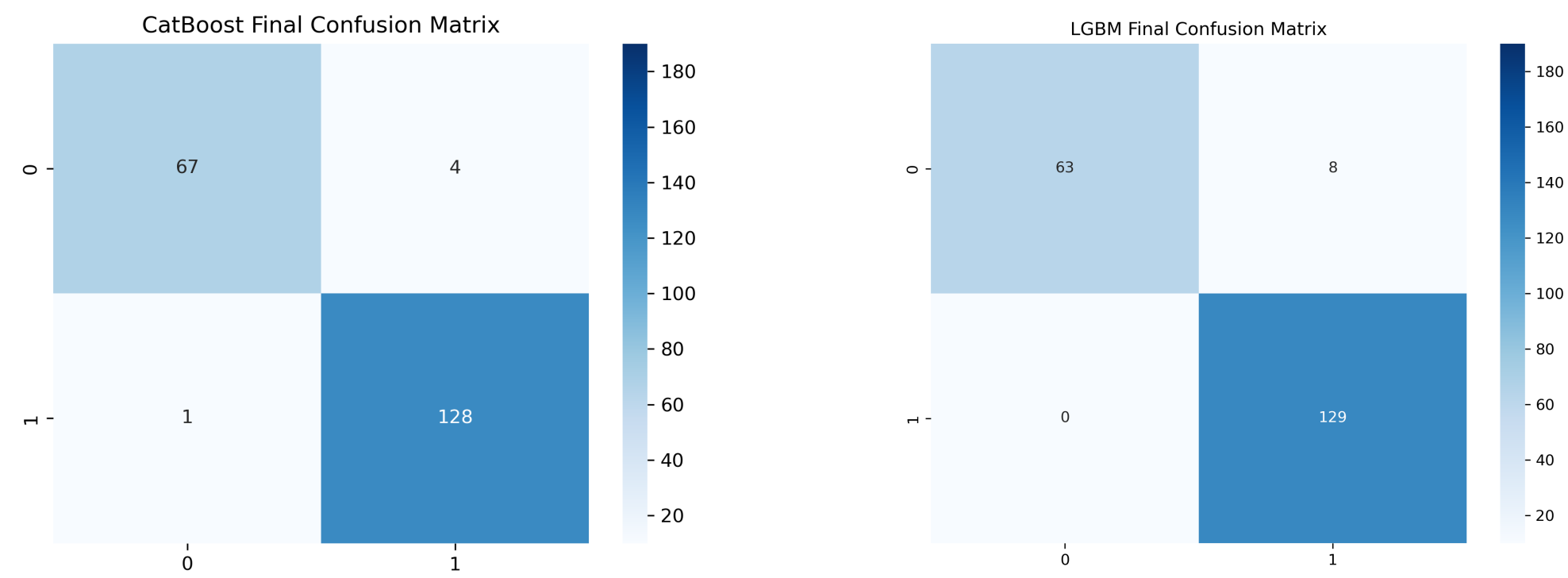

Figure 13: Confusion Matrix for CatBoost and LGBM - Final Model

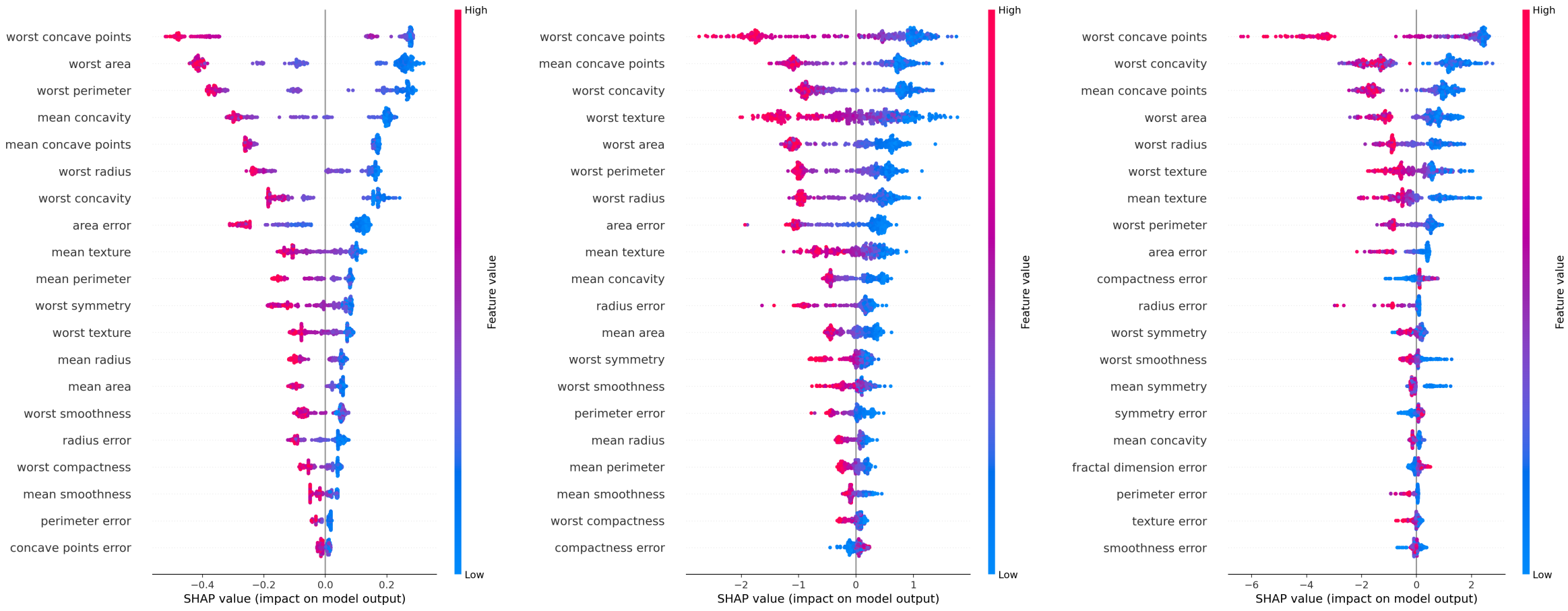

Figure 14: SHAP for XGBoost, CatBoost, and LGBM final models

# 5 Conclusion

In addition to gradient-boosting models, comparisons were made with state-of-the-art convolutional neural networks and hybrid approaches, demonstrating the competitive performance of our tuned models in the context of breast cancer detection previous studies conducted on the same dataset had obtained impressive accuracy, sometimes in 98-99% none of them focuses on reducing the False Negative, especially an optimization in a Weight $F_{beta}$. All of our final models have an improved AUC or Recall performance compared to our baseline. The AUC was greater than 99.41% and the recall 96.9%. In AdaBoost, we were able to increase the recall and reduce 25% in our False Negative. In XGBoost we were not able to increase recall performance only AUC, it was extremely hard to find a *beta* value to optimize in all four algorithms, for future work it might be suggested to optimize each algorithm separately for different values of $\beta$. For CatBoost we were able to increase the performance in AUC keeping the recall metrics the same in our baseline. LightGBM was the best model to which we were able to increase both the AUC and the recall metric, reducing the number of False Negatives. By using SHAP, we were also able to rank the final result of our models by explanatory variables, giving, in addition to a classification, it is possible to explain in each case which variable had a positive or negative impact on the final prediction.

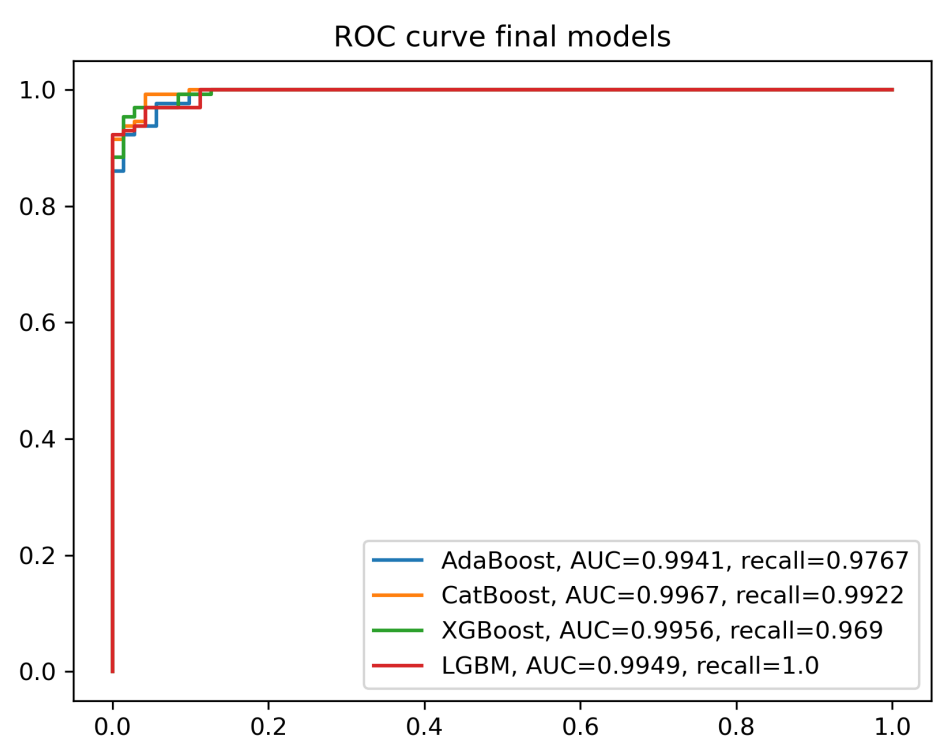

Figure 15: ROC Curve for the final models

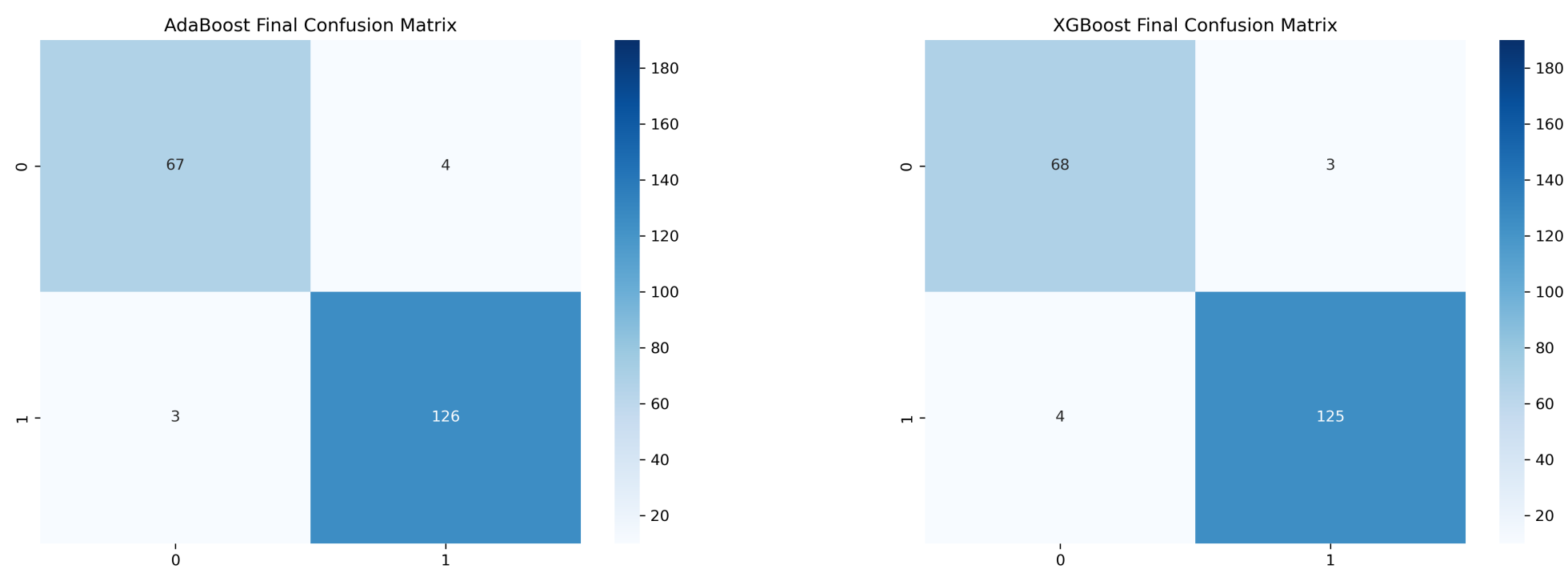

Figure 16: Confusion Matrix for AdaBoost and XGBoost - Final Model

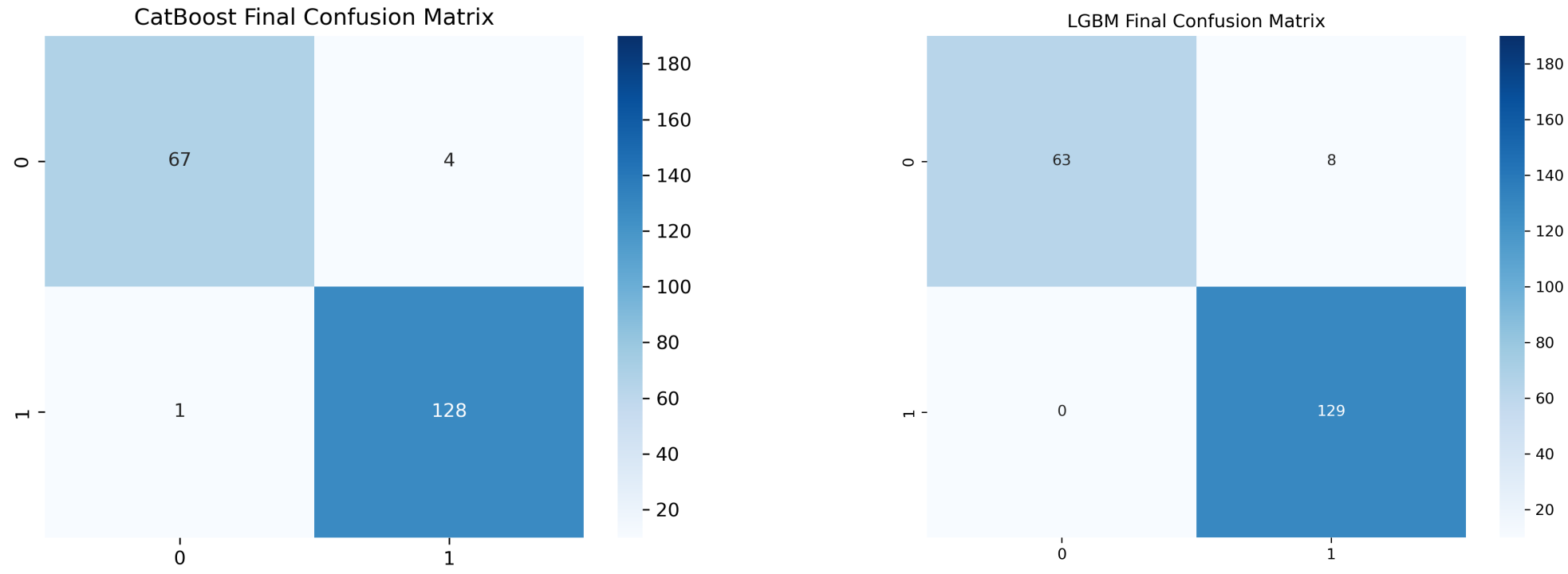

Figure 17: Confusion Matrix for CatBoost and LGBM - Final Model

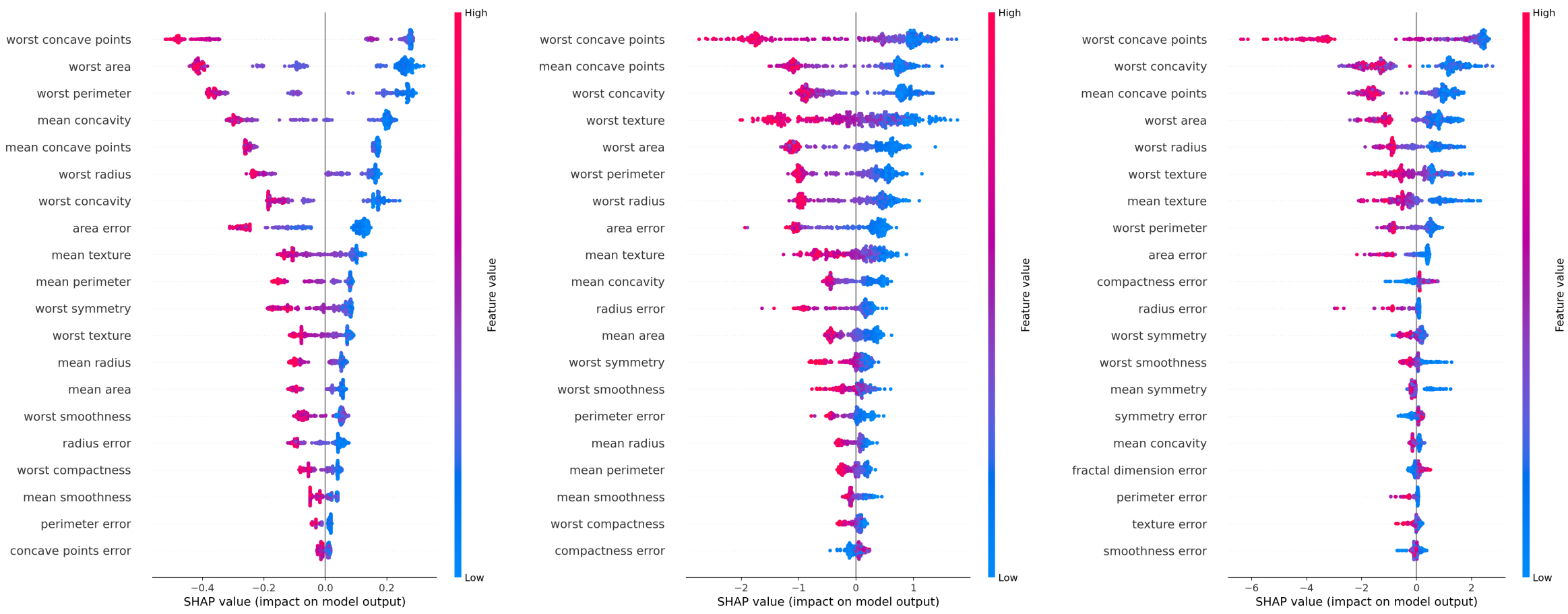

Figure 18: SHAP for XGBoost, CatBoost, and LGBM final models

We successfully determined the key features for each prediction in every instance within our dataset applying SHAP in our tree models, XGBoost, CatBoost, and LightGBM. This not only gives us an easy machine learning prediction but also ensured the model's transparency and explainability.

Future studies could explore the use of k-fold cross-validation or other advanced resampling techniques to improve the robustness of the results, particularly in handling imbalanced datasets. Furthermore, incorporating more diverse datasets and testing the models in real-world clinical settings would further validate the applicability of the proposed approach.

# 6 GitHub Source Code

https://github.com/joaomh/article-breast-cancer-classification-boosting.

# References

[1] W. H. Organization. (2023) Breast cancer. [Online]. Available: https://www.who.int/news-room/fact-sheets/detail/breast-cancer

[2] M. Arnold, E. Morgan, H. Rumgay, A. Mafra, D. Singh, M. Laversanne, J. Vignat, J. R. Gralow, F. Cardoso, S. Siesling, and I. Soerjomataram, "Current and future burden of breast cancer: Global statistics for 2020 and 2040," *The Breast*, vol. 66, pp. 15–23, 2022. [Online]. Available: https://www.sciencedirect.com/science/article/pii/S0960977622001448

[3] M. d. O. Santos, F. C. d. S. d. Lima, L. F. L. Martins, J. F. P. Oliveira, L. M. d. Almeida, and M. d. C. Cancela, "Estimativa de incidÃªncia de cÃ¢ncer no brasil, 2023-2025," *Revista Brasileira de Cancerologia*, vol. 69, no. 1, p. eâ213700, fev. 2023. [Online]. Available: https://rbc.inca.gov.br/index.php/revista/article/view/3700

[4] E. Orrantia-Borunda, P. Anchondo-Nuñez, L. E. Acuña-Aguilar, F. O. Gómez-Valles, and C. A. Ramírez-Valdespino, "Subtypes of breast cancer," in *Breast Cancer*. Exon Publications, Aug. 2022, pp. 31–42.

[5] K. S. Johnson, E. F. Conant, and M. S. Soo, "Molecular Subtypes of Breast Cancer: A Review for Breast Radiologists," *Journal of Breast Imaging*, vol. 3, no. 1, pp. 12–24, 12 2020. [Online]. Available: https://doi.org/10.1093/jbi/wbaa110

[6] H. Sung, J. Ferlay, R. L. Siegel, M. Laversanne, I. Soerjomataram, A. Jemal, and F. Bray, "Global cancer statistics 2020: Globocan estimates of incidence and mortality worldwide for 36 cancers in 185 countries," *CA: A Cancer Journal for Clinicians*, vol. 71, no. 3, pp. 209–249, 2021. [Online]. Available: https://acsjournals.onlinelibrary.wiley.com/doi/abs/10.3322/caac.21660

[7] A. Burguin, C. Diorio, and F. Durocher, "Breast cancer treatments: Updates and new challenges," *Journal of Personalized Medicine*, vol. 11, no. 8, 2021. [Online]. Available: https://www.mdpi.com/2075-4426/11/8/808

[8] S. Ivanov and L. Prokhorenkova, "Boost then convolve: Gradient boosting meets graph neural networks," *CoRR*, vol. abs/2101.08543, 2021. [Online]. Available: https://arxiv.org/abs/2101.08543

[9] S. Badirli, X. Liu, Z. Xing, A. Bhowmik, and S. S. Keerthi, "Gradient boosting neural networks: Grownet," *CoRR*, vol. abs/2002.07971, 2020. [Online]. Available: https://arxiv.org/abs/2002.07971

[10] M. T. Ribeiro, S. Singh, and C. Guestrin, ""why should i trust you?": Explaining the predictions of any classifier," in *Proceedings of the 22nd ACM SIGKDD International Conference on Knowledge Discovery and Data Mining*, ser. KDD '16. New York, NY, USA: Association for Computing Machinery, 2016, p. 1135â1144. [Online]. Available: https://doi.org/10.1145/2939672.2939778

[11] J. Sun, C.-K. Sun, Y.-X. Tang, T.-C. Liu, and C.-J. Lu, "Application of shap for explainable machine learning on age-based subgrouping mammography questionnaire data for positive mammography prediction and risk factor identification," *Healthcare*, vol. 11, no. 14, 2023. [Online]. Available: https://www.mdpi.com/2227-9032/11/14/2000

[12] L. Antwarg, C. Galed, N. Shimoni, L. Rokach, and B. Shapira, "Shapley-based feature augmentation," *Information Fusion*, vol. 96, pp. 92–102, 2023. [Online]. Available: https://www.sciencedirect.com/science/article/pii/S156625352300091X

[13] S. Kaufman, S. Rosset, and C. Perlich, "Leakage in data mining: Formulation, detection, and avoidance," vol. 6, 01 2011, pp. 556–563.

[14] G. M. M and S. P, "A survey on machine learning approaches used in breast cancer detection," in *2022 4th International Conference on Inventive Research in Computing Applications (ICIRCA)*, 2022, pp. 786–792.

[15] J. M. H. Pinheiro and M. Becker, "Um estudo sobre algoritmos de boosting e a otimizaÃ§Ã£o de hiperparÃ¢metros utilizando optuna," 2023. [Online]. Available: https://bdta.abcd.usp.br/item/003122385

[16] R. Rabiei, S. M. Ayyoubzadeh, S. Sohrabei, M. Esmaeili, and A. Atashi, "Prediction of breast cancer using machine learning approaches," *Journal of Biomedical Physics and Engineering*, vol. 12, no. 3, pp. 297–308, 2022. [Online]. Available: https://jbpe.sums.ac.ir/article_48331.html

[17] T. Pang, J. H. D. Wong, W. L. Ng, and C. S. Chan, "Deep learning radiomics in breast cancer with different modalities: Overview and future," *Expert Systems with Applications*, vol. 158, p. 113501, 2020. [Online]. Available: https://www.sciencedirect.com/science/article/pii/S0957417420303250

[18] T. Mahmood, J. Li, Y. Pei, F. Akhtar, A. Imran, and K. U. Rehman, "A brief survey on breast cancer diagnostic with deep learning schemes using multi-image modalities," *IEEE Access*, vol. 8, pp. 165 779–165 809, 2020.

[19] A. U. Haq, J. P. Li, A. Saboor, J. Khan, S. Wali, S. Ahmad, A. Ali, G. A. Khan, and W. Zhou, "Detection of breast cancer through clinical data using supervised and unsupervised feature selection techniques," *IEEE Access*, vol. 9, pp. 22 090–22 105, 2021.

[20] M. O. S. N. Wolberg, William and W. Street, "Breast Cancer Wisconsin (Diagnostic)," UCI Machine Learning Repository, 1995, DOI: https://doi.org/10.24432/C5DW2B.

[21] S. Ara, A. Das, and A. Dey, "Malignant and benign breast cancer classification using machine learning algorithms," in *2021 International Conference on Artificial Intelligence (ICAI)*, 2021, pp. 97–101.

[22] I. Ozcan, H. Aydin, and A. Ãetinkaya, "Comparison of classification success rates of different machine learning algorithms in the diagnosis of breast cancer," *Asian Pacific journal of cancer prevention : APJCP*, vol. 23, pp. 3287–3297, 10 2022.

[23] A. Khalid, A. Mehmood, A. Alabrah, B. F. Alkhamees, F. Amin, H. AlSalman, and G. S. Choi, "Breast cancer detection and prevention using machine learning," *Diagnostics*, vol. 13, no. 19, 2023. [Online]. Available: https://www.mdpi.com/2075-4418/13/19/3113

[24] M. A. Naji, S. E. Filali, K. Aarika, E. H. Benlahmar, R. A. Abdelouhahid, and O. Debauche, "Machine learning algorithms for breast cancer prediction and diagnosis," *Procedia Computer Science*, vol. 191, pp. 487–492, 2021, the 18th International Conference on Mobile Systems and Pervasive Computing (MobiSPC), The 16th International Conference on Future Networks and Communications (FNC), The 11th International Conference on Sustainable Energy Information Technology. [Online]. Available: https://www.sciencedirect.com/science/article/pii/S1877050921014629

[25] M. M. Islam, H. Iqbal, M. R. Haque, and M. K. Hasan, "Prediction of breast cancer using support vector machine and k-nearest neighbors," in *2017 IEEE Region 10 Humanitarian Technology Conference (R10-HTC)*, 2017, pp. 226–229.

[26] T. Thomas, N. Pradhan, and V. S. Dhaka, "Comparative analysis to predict breast cancer using machine learning algorithms: A survey," in *2020 International Conference on Inventive Computation Technologies (ICICT)*, 2020, pp. 192–196.

[27] Irmawati, F. Ernawan, M. Fakhreldin, and A. Saryoko, "Deep learning method based for breast cancer classification," in *2023 International Conference on Information Technology Research and Innovation (ICITRI)*, 2023, pp. 13–16.

[28] P. S. Kohli and S. Arora, "Application of machine learning in disease prediction," in *2018 4th International Conference on Computing Communication and Automation (ICCCA)*, 2018, pp. 1–4.

[29] S. Kabiraj, M. Raihan, N. Alvi, M. Afrin, L. Akter, S. A. Sohagi, and E. Podder, "Breast cancer risk prediction using xgboost and random forest algorithm," in *2020 11th International Conference on Computing, Communication and Networking Technologies (ICCCNT)*, 2020, pp. 1–4.

[30] U. Ojha and S. Goel, "A study on prediction of breast cancer recurrence using data mining techniques," in *2017 7th International Conference on Cloud Computing, Data Science & Engineering - Confluence*, 2017, pp. 527–530.

[31] A. Sharma, S. Kulshrestha, and S. Daniel, "Machine learning approaches for breast cancer diagnosis and prognosis," pp. 1–5, 12 2017.

[32] W. Wolberg, "Breast Cancer Wisconsin (Original)," UCI Machine Learning Repository, 1992, DOI: https://doi.org/10.24432/C5HP4Z.

[33] A. Bharat, N. Pooja, and R. A. Reddy, "Using machine learning algorithms for breast cancer risk prediction and diagnosis," in *2018 3rd International Conference on Circuits, Control, Communication and Computing (I4C)*, 2018, pp. 1–4.

[34] S. Das and D. Biswas, "Prediction of breast cancer using ensemble learning," in *2019 5th International Conference on Advances in Electrical Engineering (ICAEE)*, 2019, pp. 804–808.

[35] T. Islam, A. Kundu, N. Islam Khan, C. Chandra Bonik, F. Akter, and M. Jihadul Islam, “Machine learning approaches to predict breast cancer: Bangladesh perspective,” in *Ubiquitous Intelligent Systems*, P. Karuppusamy, F. P. García Márquez, and T. N. Nguyen, Eds. Singapore: Springer Nature Singapore, 2022, pp. 291–305.

[36] B. M. Abed, K. Shaker, H. A. Jalab, H. Shaker, A. M. Mansoor, A. F. Alwan, and I. S. Al-Gburi, “A hybrid classification algorithm approach for breast cancer diagnosis,” in *2016 IEEE Industrial Electronics and Applications Conference (IEACon)*, 2016, pp. 269–274.

[37] D. Borkin, A. Nemethova, G. Michalconok, and K. Maiorov, “Impact of data normalization on classification model accuracy,” *Research Papers Faculty of Materials Science and Technology Slovak University of Technology*, vol. 27, pp. 79–84, 09 2019.

[38] D. Singh and B. Singh, “Investigating the impact of data normalization on classification performance,” *Applied Soft Computing*, vol. 97, p. 105524, 2020. [Online]. Available: https://www.sciencedirect.com/science/article/pii/S1568494619302947

[39] K. Cabello-Solorzano, I. Ortigosa de Araujo, M. Peña, L. Correia, and A. J. Tallón-Ballesteros, “The impact ofÂ data normalization onÂ theÂ accuracy ofÂ machine learning algorithms: A comparative analysis,” in *18th International Conference on Soft Computing Models in Industrial and Environmental Applications (SOCO 2023)*, P. García Bringas, H. Pérez García, F. J. Martínez de Pisón, F. Martínez Álvarez, A. Troncoso Lora, Á. Herrero, J. L. Calvo Rolle, H. Quintián, and E. Corchado, Eds. Cham: Springer Nature Switzerland, 2023, pp. 344–353.

[40] T. Hastie, R. Tibshirani, and J. Friedman, *The Elements of Statistical Learning*, 2nd ed. Springer New York, NY, 2009.

[41] R. E. Schapire, “A brief introduction to boosting,” in *Proceedings of the 16th International Joint Conference on Artificial Intelligence - Volume 2*, ser. IJCAI’99. San Francisco, CA, USA: Morgan Kaufmann Publishers Inc., 1999, p. 1401â1406.

[42] J. H. Friedman, “Greedy function approximation: A gradient boosting machine.” *The Annals of Statistics*, vol. 29, no. 5, pp. 1189 – 1232, 2001. [Online]. Available: https://doi.org/10.1214/aos/1013203451

[43] S. R. E. and F. Yoav, “A brief introduction to boosting,” in *Proceedings of the 16th International Joint Conference on Artificial Intelligence - Volume 2*, ser. IJCAI’99. San Francisco, CA, USA: Morgan Kaufmann Publishers Inc., 1999, p. 1401â1406.

[44] T. Chen and C. Guestrin, “Xgboost: A scalable tree boosting system,” *CoRR*, vol. abs/1603.02754, 2016. [Online]. Available: http://arxiv.org/abs/1603.02754

[45] A. V. Dorogush, A. Gulin, G. Gusev, N. Kazeev, L. O. Prokhorenkova, and A. Vorobev, “Fighting biases with dynamic boosting,” *CoRR*, vol. abs/1706.09516, 2017. [Online]. Available: http://arxiv.org/abs/1706.09516

[46] G. Ke, Q. Meng, T. Finley, T. Wang, W. Chen, W. Ma, Q. Ye, and T.-Y. Liu, “Lightgbm: A highly efficient gradient boosting decision tree,” in *Advances in Neural Information Processing Systems*, I. Guyon, U. V. Luxburg, S. Bengio, H. Wallach, R. Fergus, S. Vishwanathan, and R. Garnett, Eds., vol. 30. Curran Associates, Inc., 2017. [Online]. Available: https://proceedings.neurips.cc/paper/2017/file/6449f44a102fde848669bdd9eb6b76fa-Paper.pdf

[47] M. Kuhn and K. Johnson, *Applied Predictive Modeling*, 1st ed. Springer New York, NY, 2013.

[48] T. Fawcett, “An introduction to roc analysis,” *Pattern Recognition Letters*, vol. 27, no. 8, pp. 861–874, 2006, rOC Analysis in Pattern Recognition. [Online]. Available: https://www.sciencedirect.com/science/article/pii/S016786550500303X

[49] C. D. Brown and H. T. Davis, "Receiver operating characteristics curves and related decision measures: A tutorial," *Chemometrics and Intelligent Laboratory Systems*, vol. 80, no. 1, pp. 24–38, 2006. [Online]. Available: https://www.sciencedirect.com/science/article/pii/S0169743905000766

[50] J. Hayes, A. Dekhtyar, and S. Sundaram, "Advancing candidate link generation for requirements tracing: the study of methods," *IEEE Transactions on Software Engineering*, vol. 32, no. 1, pp. 4–19, 2006.

[51] T. Merten, D. Krämer, B. Mager, P. Schell, S. Bürsner, and B. Paech, "Do information retrieval algorithms for automated traceability perform effectively on issue tracking system data?" in *Requirements Engineering: Foundation for Software Quality*, M. Daneva and O. Pastor, Eds. Cham: Springer International Publishing, 2016, pp. 45–62.

[52] D. M. Berry, "Evaluation of tools for hairy requirements and software engineering tasks," in *2017 IEEE 25th International Requirements Engineering Conference Workshops (REW)*, 2017, pp. 284–291.

[53] J. P. Winkler, J. GrÃ¶nberg, and A. Vogelsang, "Optimizing for recall in automatic requirements classification: An empirical study," in *2019 IEEE 27th International Requirements Engineering Conference (RE)*, 2019, pp. 40–50.

[54] T. Akiba, S. Sano, T. Yanase, T. Ohta, and M. Koyama, "Optuna: A next-generation hyperparameter optimization framework," *CoRR*, vol. abs/1907.10902, 2019. [Online]. Available: http://arxiv.org/abs/1907.10902

[55] L. S. Shapley, *17. A Value for n-Person Games*. Princeton: Princeton University Press, 1953, pp. 307–318. [Online]. Available: https://doi.org/10.1515/9781400881970-018

[56] S. M. Lundberg, G. Erion, H. Chen, A. DeGrave, J. M. Prutkin, B. Nair, R. Katz, J. Himmelfarb, N. Bansal, and S.-I. Lee, "From local explanations to global understanding with explainable ai for trees," *Nature Machine Intelligence*, vol. 2, no. 1, pp. 2522–5839, 2020.

[57] S. M. Lundberg and S.-I. Lee, "Shap documentation." [Online]. Available: https://shap.readthedocs.io/en/latest/

[58] S. M. Lundberg and S. Lee, "A unified approach to interpreting model predictions," *CoRR*, vol. abs/1705.07874, 2017. [Online]. Available: http://arxiv.org/abs/1705.07874